\pdfoutput=1

\documentclass[11pt]{article}

\usepackage[]{ACL2023}

\usepackage{times}
\usepackage{latexsym}

\usepackage[T1]{fontenc}

\usepackage[utf8]{inputenc}

\usepackage{microtype}

\usepackage{inconsolata}

\usepackage{listings}
\usepackage{multirow}
\usepackage{graphicx}
\usepackage{colortbl}
\usepackage{booktabs}
\usepackage{xcolor}

\title{NumeroLogic: Number Encoding \\for Enhanced LLMs' Numerical Reasoning}

\author{
Eli Schwartz$^{1}$, Leshem Choshen$^{2,3}$, Joseph Shtok$^{1}$, \\
\textbf{Sivan Doveh$^{1}$, Leonid Karlinsky$^{2}$, Assaf Arbelle$^{1}$} \\
\\
$^{1}$IBM Research, $^{2}$MIT-IBM Watson AI Lab, $^{3}$MIT
}

\begin{document}
\maketitle

\begin{abstract}
Language models struggle with handling numerical data and performing arithmetic operations.
We hypothesize that this limitation can be partially attributed to non-intuitive textual numbers representation.
When a digit is read or generated by a causal language model it does not know its place value (e.g. thousands vs. hundreds) until the entire number is processed.
To address this issue, we propose a simple adjustment to how numbers are represented by including the count of digits before each number. For instance, instead of \texttt{"42"}, we suggest using \texttt{"{2:42}"} as the new format. This approach, which we term NumeroLogic, offers an added advantage in number generation by serving as a Chain of Thought (CoT). By requiring the model to consider the number of digits first, it enhances the reasoning process before generating the actual number.
We use arithmetic tasks to demonstrate the effectiveness of the NumeroLogic formatting. We further demonstrate NumeroLogic applicability to general natural language modeling, improving language understanding performance in the MMLU benchmark.
\end{abstract}

\section{Introduction}
\vspace{-5pt}

Large Language Models (LLMs) struggle with numerical and arithmetical tasks.
Despite continuous improvements, even the most advanced models like GPT-4 \cite{achiam2023gpt} still exhibit poor performance when confronted with tasks such as multiplying 3-digit numbers \cite{shen2023positional}.
Recent studies (\cite{lee2024teaching,shen2023positional}) have proposed techniques to improve arithmetic in LLMs, such as the Chain of Thought (CoT; \cite{wei2022chain}) method, which pushes the model to anticipate the entire sequence of algorithmic steps rather than just the final output.
While these strategies offer valuable insights into the capabilities of LLMs, they primarily concentrate on post-hoc solutions for specific arithmetic challenges and do not present a practical solution for pretraining LLMs. Our research, however, focuses on solutions applicable to self-supervised language modeling in general, utilizing arithmetic exercises primarily for evaluating their impact.

We hypothesize that one of the challenges LLMs face when dealing with numerical tasks is the textual representation of numbers.
In today's most popular decoder-based LLMs, each token attends only to previous tokens.
When a model ``reads" a token representing a digit (or multiple digits) it cannot tell its place value, i.e. `1' can represent \textit{1 million}, \textit{1 thousand}, or a single unit.
Only when reaching the end of the number might the model update its representation of the previous digit tokens to be related to their real place value.

\begin{figure}
    \centering
    \includegraphics[width=0.75\columnwidth]{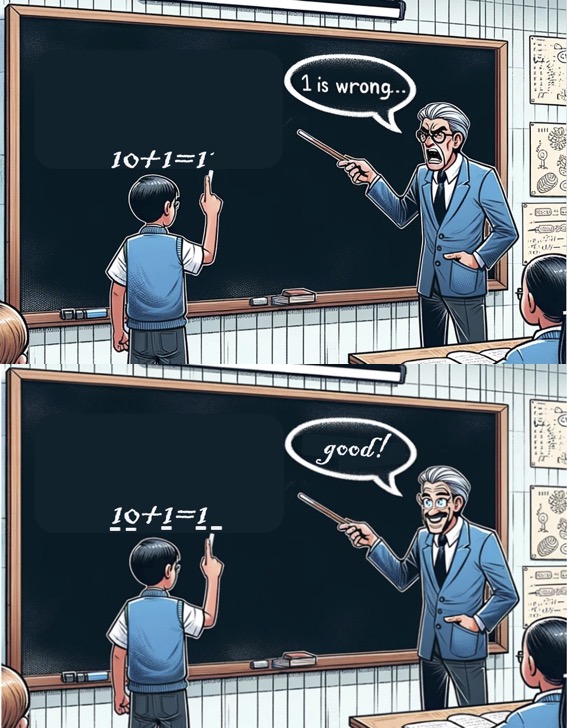}
    \caption{Reading numbers in a causal manner from left to right is sub-optimal for LLMs, as it is for humans. The model has to reach the final digits of a number before it can infer the place value of the first digit. To address this, we propose ``NumeroLogic", a numerical format where digit count is indicated before the actual number. Image by DALL-E 3 \cite{betker2023improving}.}
    \vspace{-20pt}
    \label{fig:enter-label}
\end{figure}

To address this issue, we propose a straightforward reformatting technique called "NumeroLogic," which involves adding the number of digits as a prefix to numbers.
This lets the model know in advance what is the place value of a digit before it is read. This simple change also offers another benefit, when the model is generating a number it needs to first reason about what is going to be the number of digits.
This acts as a Chain of Thought (CoT) \cite{wei2022chain}, encouraging the model to perform some reasoning before it begins to predict digits.
Implementing the suggested reformatting does not necessitate any alterations to the model's architecture; it can be accomplished through text pre- and post-processing based on regex.

We demonstrate that NumeroLogic enhances the numerical abilities of LLMs across both small and larger models (up to 7B parameters). This enhancement is showcased through supervised training on arithmetic tasks and its application in self-supervised causal language modeling to enhance general language comprehension.

\vspace{-5pt}
\section{Related Work}
\vspace{-5pt}
Recently, there has been a significant interest in enhancing the numerical capabilities of LLMs. One approach to investigating these capabilities is by assessing their performance in arithmetic tasks. Several recent studies have proposed methods to enhance performance in these tasks.
One strategy involves reversing the expected result order from the least to the most significant digit \cite{lee2024teaching}. Another strategy is using an elaborated CoT where the model is taught to predict all steps of an algorithm predefined for each arithmetic task \cite{lee2024teaching}. In \cite{shen2023positional}, it is noted that the model learns to rely too heavily on positional encoding when trained for a specific arithmetic task. They suggest ways to overcome it, e.g. adding random white spaces in the middle of numbers.
These studies aim to enhance the performance of arithmetic tasks by offering tailored solutions to the associated challenges.
In contrast, our focus is on identifying solutions that benefit general language modeling rather than just arithmetic tasks, with arithmetic tasks being used solely for measuring improvements.

Another aspect important for LLMs numerical capabilities is the tokenization process.
The commonly used Byte Pair Encoding  (BPE) based methods \cite{gage1994new,sennrich2015neural} for tokenization are based on the corpus distribution and can split a number to tokens in unintuitive ways. Different foundation models took different approaches when dealing with number tokenization. PaLM \cite{chowdhery2023palm}, Llama \cite{touvron2023llama}, and Mistral \cite{jiang2023mistral} force each digit to have a single token. GPT-3.5 and GPT-4 define a token for each up to 3-digit number \cite{achiam2023gpt}. Somewhat related to our work, in \cite{singh2024tokenization}, they highlighted an issue with the GPT approach. They show that dividing large numbers into 3-digit segments from left to right undermines arithmetic performance. They suggest overcoming it by inserting commas between digits to control the splitting. Another related work, is \cite{kim-etal-2021-seen}. They focus on the extrapolation ability of LLMs to unseen numbers and use a special number encoding that lets the LLM know the digit place-value.

\vspace{-5pt}
\section{NumeroLogic}
\vspace{-5pt}
We introduce NumeroLogic, a technique for boosting causal LLM's numerical capabilities.
The concept involves adding a digit count before numbers, enabling the model to know the place values of digits before reaching the final digits of a number.
Additionally, the model needs to predict the total number of digits before generating a number, acting as a simplified CoT, prompting it to reason about the number that is going to be generated.

We add special tokens to help represent numbers with the number-of-digit prefix, \texttt{"<startnumber>"}, \texttt{"<midnumber>"}, and \texttt{"<endnumber>"} (or, for simplicity, \texttt{"<sn>"}, \texttt{"<mn>"}, and \texttt{"<en>"}). For floating points, the prefix includes both the number of digits of the integer part and the decimal part. For example, \texttt{"42"} is replaced by \texttt{"<sn>2<mn>42<en>"} and \texttt{"3.14"} is replaced by \texttt{"<sn>1.2<mn>3.14<en>"}.
When using the LLM to generate numbers, we disregard the information about the number of digits and only retain the generated number itself.
Although not within the scope of this study, it may be feasible to leverage the additional information to identify discrepancies, wherein the model predicts a certain digit count but produces a number with a different count of digits.

For small transformers, we train all parameters from scratch with character-level tokenization. For small transformers, we also replace the special tokens with single characters, \texttt{"<sn>"}, \texttt{"<mn>"}, and \texttt{"<en>"} are replaced with \texttt{"\{"}, \texttt{":"}, and \texttt{"\}"}, respectively.
For larger transformers, we start from pre-trained models.
We add the new special tokens to the tokenizer's vocabulary and expand the embedding layer and the final fully connected layer to fit the new vocabulary size.
When continuing training on causal language modeling or fine-tuning on supervised arithmetic tasks, we use low-rank adaptation (LoRA) \cite{hu2021lora}. We apply LoRA for the attention block projection matrices (Q, K, V, O) and train the modified embedding layer and the final fully-connected layer in full rank.

The NumeroLogic approach includes basic text pre-processing and post-processing steps that occur before and after the tokenizer's encoding and decoding methods, respectively.
Both can be implemented based on regular expressions:
\vspace{-5pt}


\lstset{basicstyle=\small\ttfamily}
\begin{lstlisting}[language=python]
def preprocess_all_numbers(text):
  def f(match):
    num = match.group(0)
    i = match.group(1)
    li = len(i)
    d = match.group(3)
    ld = len(d) if d else 0
    if d:
      prefix = f'<sn>{li}.{ld}<mn>'
    else:
      prefix = f'<sn>{li}<mn>'
    return prefix + num + '<en>'
  pattern = '(\d+)(\.(\d+))?'
  return re.sub(pattern, f, text)

def postprocess_all_numbers(text):
  pattern = '<sn>[\d\.]+<mn>'
  text = re.sub(pattern, '', text)
  text = re.sub('<en>', '', text)
  return text
\end{lstlisting}

\vspace{-5pt}
\section{Experiments}
\vspace{-5pt}
To test the effect of NumeroLogic we conducted several experiments. First, we tested supervised training of a small language model (NanoGPT) on various arithmetic tasks. We then test the scalability to larger models (Llama2-7B). Finally, we test self-supervised pretraining of Llama2-7B, with the suggested formatting, and test on general language understanding tasks.

\vspace{-5pt}
\subsection{Arithmetic tasks with small model}
\vspace{-5pt}
\label{exp_nanogpt}
We trained NanoGPT \cite{NanoGPT} from scratch in a supervised manner jointly on 5 arithmetic tasks: addition, subtraction, multiplication, sine, and square root. Addition and subtraction are performed with up to 3-digit integer operands. Multiplications are performed with up to 2-digit integer operands. Sine and square root with 4 decimal-places floating point operands and results. The operand range for sine is within $\left[-\pi/2,\pi/2\right]$. The operand range for the square root is within $\left[0,10\right]$. The model is trained in a multi-task fashion on all 5 tasks, with 10K training samples for each task except for multiplication, for which 3K samples are used.
We followed the protocol from Section D.2 in \cite{lee2024teaching}.

Tab.~\ref{tab:nanogpt} compares the results of training with plain numbers to training with the NumeroLogic encoding. For addition and subtraction, a model trained with plain numbers reached $88.37\%$ and $73.76\%$ accuracy, respectively, while with the NumeroLogic encoding, the tasks are almost solved ($99.96\%$ and $97.2\%$). For multiplication, we observe more than doubling of the accuracy, from $13.81\%$ to $28.94\%$. Furthermore, for the floating point operations, sine and square root, we see a significant improvement of $~4\%$ for both tasks.


\begin{table}
    \centering
    \footnotesize
    \begin{tabular}{ccc|cc|c}
    \toprule
         & Num.  & int/ & & Numero \\
         Op. & digit & float &   Plain & Logic & Gain\\
         \midrule
+ & 3 & int & 88.37 & 99.96 & \cellcolor{green!25}+11.6 \\
$-$ & 3 & int & 73.76 & 97.20 & \cellcolor{green!25}+23.4 \\
$*$ & 2 & int & 13.81 & 28.94 & \cellcolor{green!25}+15.1 \\
sine & 4 & float & 30.59 & 34.59 & \cellcolor{green!25}+4.00 \\
sqrt & 4 & float & 22.13 & 26.66 & \cellcolor{green!25}+4.53 \\
\bottomrule
    \end{tabular}
    \vspace{-8pt}
    \caption{\textbf{NanoGPT arithmetic tasks accuracy with NumeroLogic encoding.} A single model is jointly trained for all tasks. The encoding produces high accuracy gains for all tasks.}
    \vspace{-15pt}
    \label{tab:nanogpt}
\end{table}

\vspace{-5pt}
\subsection{Arithmetic tasks with larger model}
\vspace{-5pt}

Next, we test how the method scales to a larger model. For this experiment, we fine-tune a pretrained Llama2-7B model \cite{touvron2023llama}. In this experiment, we again tested the same five arithmetic tasks: addition, subtraction, multiplication, sine, and square root.
For addition (5 digit), subtraction (5 digit), and multiplication (3 digit) we tested on two versions - integers and floating point numbers. For generating a random N-digit floating point operand we first sample an up to N-digit integer and then divide it by a denominator uniformly sampled from $\left\{10^0,10^1,...,10^N\right\}$. For each of the addition, subtraction, and multiplication tasks, we generated 300K random equations as a training set. The sine and square root operands and results are generated with 5 decimal place accuracy, we generated 30K random equations for the training sets of these tasks. Since we are working with a pretrained model we add new tokens (\texttt{"<sn>"}, \texttt{"<mn>"}, and \texttt{"<en>"}) to the tokenizer's vocabulary. We finetune one model per task with LoRA \cite{hu2021lora} (rank 8), we also train in full-rank the embedding layer and the final linear layer since they are extended to fit the larger vocab. size.

The results are presented in Tab.~\ref{tab:llama}. Addition and subtraction of integers are mostly solved by a model as large as Llama2-7B even for much larger numbers (e.g. 20-digit). For our 5-digit experiments, the plain text baselines reached 99.86\% and 99.6\% performance, for addition and subtraction, respectively. Despite the high performance of plain text, we still observe an improvement when using NumerLogic, with a perfect 100\% for addition and rectification of more than 80\%  of the subtraction mistakes, reaching 99.93\% accuracy for subtraction.
For all other, non-saturated, tasks we observed significant gains of 1\%-6\%.


\begin{table}
    \centering
    \footnotesize
    \begin{tabular}{ccc|cc|c}
    \toprule
         & Num.  & int/ & & Numero \\
         Op. & digit & float &   Plain & Logic & Gain\\
         \midrule
+ & 5 & int & 99.86 & 100.0 & \cellcolor{green!25}+0.14 \\
$-$ & 5 & int & 99.60 & 99.93 & \cellcolor{green!25}+0.33 \\
$*$ & 3 & int & 34.20 & 35.33 & \cellcolor{green!25}+1.13 \\
+ & 5 & float & 91.40 & 94.43 & \cellcolor{green!25}+3.03 \\
$-$ & 5 & float & 88.76 & 92.73 & \cellcolor{green!25}+3.97 \\
$*$ & 3 & float & 24.73 & 31.03 & \cellcolor{green!25}+6.30 \\
sine & 5 & float & 25.06 & 28.13 & \cellcolor{green!25}+3.07 \\
sqrt & 5 & float & 13.00 & 17.16 & \cellcolor{green!25}+4.16 \\
\bottomrule
    \end{tabular}
    \vspace{-5pt}
    \caption{\textbf{Llama2-7B arithmetic tasks accuracy with NumeroLogic encoding.} We observe significant gains thanks to the NuemroLogic encoding for all tasks where performance is not saturated.}
    \vspace{-15pt}
    \label{tab:llama}
\end{table}

\vspace{-5pt}
\subsection{Self-Supervised Pretraining}
\vspace{-5pt}
Our approach differs from other methods in that it is not specialized for a specific task, such as arithmetic, but rather designed for general language modeling tasks involving text with numerical values.
To test this capability we continue the pretraining of LLama2-7B with the causal text modeling objective (next token prediction).
We train on text from the RefinedWeb dataset \cite{refinedweb}.
The goal is to teach the model to read and write numbers in the NumeroLogic format without forgetting its previously acquired knowledge. To facilitate this, we perform the continued pretraining with LoRA.
We then test the model in a 0-shot manner on MMLU \cite{hendryckstest2021,hendrycks2021ethics}.

\begin{figure}
    \vspace{-10pt}
    \hspace{-20pt}
    \includegraphics[width=1.1\linewidth]{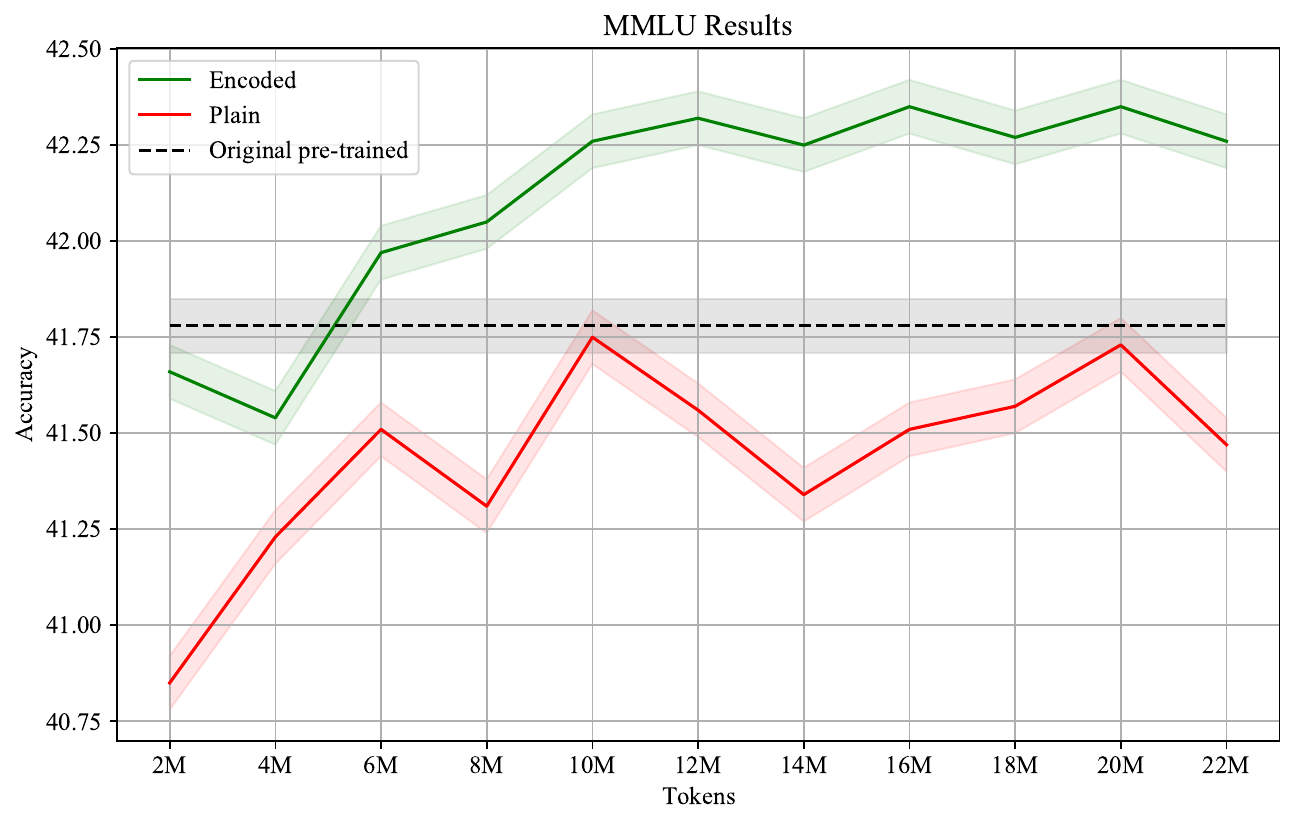}
    \vspace{-25pt}
    \caption{MMLU Accuracy of Llama2-7B. Continuing self-supervised pretraining on web-curated text tokens, when numbers are encoded with NumeroLogic, helps improve the performance beyond the pretrained model or a model trained on the same text with plain numbers.}
    \label{fig:mmlu}
\end{figure}

In Fig.~\ref{fig:mmlu}, we present the MMLU 0-shot results obtained from training the model using plain numbers versus NumeroLogic encoding on an equal number of tokens.
While training with plain numbers does not enhance the model's accuracy compared to the pretrained model, employing NumeroLogic encoding results in a statistically significant improvement of 0.5\%.
The MMLU benchmark encompasses tasks from diverse domains, some emphasizing analytical skills and numerical comprehension while others do not. In Tab. \ref{tab:mmlu_by_cat}, we delve into the impact of NumeroLogic on MMLU tasks categorized by field. As anticipated, tasks in STEM fields exhibit more substantial enhancements compared to those in social sciences and humanities. Tab. \ref{tab:with_numbers_vs_wo} provides a detailed analysis of NumeroLogic's performance boost across tasks containing numbers versus those that do not. Consistently, tasks involving numbers show higher improvement.

\begin{table}[h]
    \centering
    \vspace{-8pt}
    \footnotesize
    \begin{tabular}{lc}
    \toprule
         & Change\\
         \midrule
         Social sciences & +0.1\%\\
         Humanities & +0.43\%\\
         STEM & +0.79\%\\
         Others & +1.19\%\\
         \bottomrule
    \end{tabular}
    \vspace{-8pt}
    \caption{MMLU accuracy change due to NumeroLogic encoding on tasks from different fields. STEM tasks which are more likely to require numerical understanding enjoy higher improvement.}
    \vspace{-15pt}
    \label{tab:mmlu_by_cat}
\end{table}

\begin{table}[h]
    \centering
    \footnotesize
    \begin{tabular}{lc}
    \toprule
         & Change\\
         \midrule
         Tasks with numbers & +1.16\%\\
         Tasks without numbers & +0.14\%\\
         \bottomrule
    \end{tabular}
    \vspace{-8pt}
    \caption{MMLU accuracy change due to NumeroLogic encoding on tasks with and without numbers. Tasks with numbers enjoy higher improvement.}
    \label{tab:with_numbers_vs_wo}
\end{table}

\vspace{-5pt}
\subsection{Ablation studies}
\vspace{-5pt}

\subsubsection{Encoding operands vs. results}
We experimented to test the effect of operand encoding vs. the expected output (equation result) encoding. Operand encoding primarily influences the model's comprehension of numerical values in the input, while result encoding is more associated with CoT, prompting the model first reason about the expected number of digits.
We repeat the experiment from Section~\ref{exp_nanogpt}, but with the NumeroLogic encoding applied only to the operands or to the results and report the 3-digit addition results for the different variants. The results are presented in Tab.~\ref{tab:operands_vs_results}. We find that both operands and results encodings are beneficial, with a stronger impact attributed to encoding the results. Applying NumeroLogic to all numbers, both operands and results, yields the highest level of accuracy.

\vspace{-5pt}
\subsubsection{Different Encodings}
\vspace{-5pt}
We experimented with different formats for providing the number of digits. One alternative we tested is defining a set of new special tokens representing each possible number of digits, \{\texttt{<1digitnumber>, <2digitnumber>,...}\}. We observed that the performance of having multiple special tokens is even lower than plain numbers. It might be due to the unbalanced distribution of numbers. E.g. numbers with a single digit are much less frequent in the data of 3-digit additions, it is possible the model has not seen enough single-digit numbers to learn a good representation of the \texttt{<1digitnumber>} token.
Another alternative we tested is removing the ``end of number" token (\texttt{<en>}), keeping only the number prefix, e.g. \texttt{"<sn>3<mn>100"}. This works better than plain but slightly worse than the full NumeroLogic encoding.
The results are summarized in Tab.~\ref{tab:enc_ablation}.

\begin{table}
    \centering
    \small
    \begin{tabular}{cccc}
    \toprule
    &&\multicolumn{2}{c}{Operands}\\
         \multirow{2}{*}{\rotatebox[origin=rb]{90}{Result}} &  & Plain & Encoded \\
           & Plain &  \cellcolor{red!70!white} 88.37\% & \cellcolor{orange!60!white} 98.05\%\\
         & Encoded & \cellcolor{orange} 89.34\% & \cellcolor{green} 99.78\%\\
         \bottomrule
    \end{tabular}
    \vspace{-5pt}
    \caption{Testing the effect of encoding the equation's operands vs. result. Tested on the addition task with NanoGPT. Either encoding the operands (i.e. input comprehension) or encoding the results (i.e. CoT effect) have a positive effect, with a stronger effect for operands' encoding. Encoding both the operands and the result provides the best performance.}
    \label{tab:operands_vs_results}
\end{table}

\begin{table}[t]
    \centering
    \small
    \begin{tabular}{lc}
    \toprule
         Encoding &  Accuracy\\
         \midrule
         Plain (e.g. \texttt{"100"}) & 34.20\%\\
         Multi special tokens (\texttt{"<3digitnumber>100"}) & 33.56\%\\
         Only prefix (\texttt{"<sn>3<mn>100"}) & 34.93\%\\
         NumeroLogic (\texttt{"<sn>3<mn>100<en>"}) & 35.33\%\\
    \bottomrule
    \end{tabular}
    \vspace{-5pt}
    \caption{\textbf{Different encoding alternatives} performance on 3-digit integer multiplications.}
    \label{tab:enc_ablation}
\end{table}

\vspace{-5pt}
\subsubsection{Is it the extra tokens?}
\vspace{-5pt}
It has been shown that the advantage of CoT is at least partially due to the extra tokens that allow the model to perform more computations or store information \cite{pfau2024let}.
To understand the effect of the extra tokens we run an experiment where all the extra tokens introduced by NumeroLogic are replaced with filler white-space tokens.
Additionally, in \cite{shen2023positional}, it has been shown that the model learns to rely too heavily on positional encoding when trained on arithmetic tasks. It causes failures when the model is tested with numbers less frequent in the training data (e.g. 1 1-digit numbers when the model is trained on up to 3-digit numbers).
To deal with this limitation, they suggest adding filler white-space tokens at random locations between the digits.
We also report the results of their approach \cite{shen2023positional} where we use the same number of tokens as NumeroLogic would have required, just that they are replaced with white-space tokens at random locations.
These experiments were performed by finetuning Llama2-7B on 3-digit floating-point multiplication.
The results are reported in Table~\ref{tab:extra_tokens}.
We observe that just adding the extra tokens does not help and the performance is similar to the plain format.
Adding the same amount of extra tokens in random locations is somewhat helpful but not as effective as NumeroLogic. It eliminates the model's reliance on positional encoding but does not provide place-value information like NumeroLogic.

\begin{table}[]
    \centering
    \small
    \begin{tabular}{llc}
    \toprule
        Format & Example & Accuracy \\
        \midrule
        NumeroLogic & \texttt{\{1:1\}*\{1:1\}=\{1:1\}} & 31.03\% \\
        White-spaces & \texttt{\_\_\_1\_*\_\_\_1\_=\_\_\_1\_} & 24.37\% \\
        Random white-spaces & \texttt{\_\_\_\_1*\_\_1\_\_=1\_\_\_\_} & 27.76\% \\
        Plain & \texttt{1*1=1} & 24.73\% \\
         \bottomrule
    \end{tabular}
    \vspace{-7pt}
    \caption{\textbf{Extra tokens effect:} Just adding filler white space tokens is not helpful and is comparable to the plain format. The random white-space method \cite{shen2023positional} of adding filler tokens at random locations is helpful but less effective compared to NumeroLogic.}
    \label{tab:extra_tokens}
\end{table}

\vspace{-7pt}
\section{Conclusions}
\vspace{-7pt}

We introduced NumeroLogic, a novel method to improve language models' handling of numerical data. Our approach prefixes numbers with their digit count, enhancing models' understanding of place value and prompting better reasoning about numbers' magnitude, akin to chain-of-thought reasoning.
We tested NumeroLogic on both arithmetic and broader language understanding tasks. The results showed substantial enhancements in numerical tasks, including integer and floating-point calculations, and in broader modeling contexts like the MMLU benchmark.
In summary, NumeroLogic is a straightforward yet effective enhancement for language models' numerical abilities, applicable across various tasks without requiring changes to the models' architecture.

\section{Limitations}

The NumeroLogic encoding, while enhancing numerical reasoning, might increase the number of tokens per number. Moreover, it introduces additional steps in pre- and post-processing. This raises the computational costs and also potentially increases the model's latency during inference. These factors might impact the efficiency of NumeroLogic, especially in numerical-processing-intensive applications.

Our experiments predominantly involved fine-tuning pre-trained language models (LLMs) rather than training them from scratch with NumeroLogic. While this limits our ability to conclusively predict the impacts from the pre-training phase, incorporating NumeroLogic early in the pre-training would likely have a stronger positive rather than negative effect on the performance. Additionally, our testing did not extend to models larger than 7B parameters. However,  it has been demonstrated that both small and large models exhibit similar learning behaviors \cite{warstadt2023findings}; therefore, it is plausible to predict that scaling up the model size will not diminish the effectiveness of NumeroLogic.

Lastly, our evaluation was confined to controlled academic benchmarks, which might not fully represent the complexities of real-world numerical data. Extending testing to diverse, real-world datasets is essential to fully understand NumeroLogic's practical effectiveness and ensure it can handle the unpredictable nature of real-world numerical data.
Similarly, despite caring mainly about numerical aspects, we checked English-focused datasets and data. The cross effects with different languages, scripts and even numerical writing system is left as an open question.

\bibliography{anthology,bib}

\begin{thebibliography}{19}
\expandafter\ifx\csname natexlab\endcsname\relax\def\natexlab#1{#1}\fi

\bibitem[{Achiam et~al.(2023)Achiam, Adler, Agarwal, Ahmad, Akkaya, Aleman, Almeida, Altenschmidt, Altman, Anadkat et~al.}]{achiam2023gpt}
Josh Achiam, Steven Adler, Sandhini Agarwal, Lama Ahmad, Ilge Akkaya, Florencia~Leoni Aleman, Diogo Almeida, Janko Altenschmidt, Sam Altman, Shyamal Anadkat, et~al. 2023.
\newblock Gpt-4 technical report.
\newblock \emph{arXiv preprint arXiv:2303.08774}.

\bibitem[{Betker et~al.(2023)Betker, Goh, Jing, Brooks, Wang, Li, Ouyang, Zhuang, Lee, Guo et~al.}]{betker2023improving}
James Betker, Gabriel Goh, Li~Jing, Tim Brooks, Jianfeng Wang, Linjie Li, Long Ouyang, Juntang Zhuang, Joyce Lee, Yufei Guo, et~al. 2023.
\newblock Improving image generation with better captions.
\newblock \emph{Computer Science. https://cdn. openai. com/papers/dall-e-3. pdf}, 2(3):8.

\bibitem[{Chowdhery et~al.(2023)Chowdhery, Narang, Devlin, Bosma, Mishra, Roberts, Barham, Chung, Sutton, Gehrmann et~al.}]{chowdhery2023palm}
Aakanksha Chowdhery, Sharan Narang, Jacob Devlin, Maarten Bosma, Gaurav Mishra, Adam Roberts, Paul Barham, Hyung~Won Chung, Charles Sutton, Sebastian Gehrmann, et~al. 2023.
\newblock Palm: Scaling language modeling with pathways.
\newblock \emph{Journal of Machine Learning Research}, 24(240):1--113.

\bibitem[{Gage(1994)}]{gage1994new}
Philip Gage. 1994.
\newblock A new algorithm for data compression.
\newblock \emph{The C Users Journal}, 12(2):23--38.

\bibitem[{Hendrycks et~al.(2021{\natexlab{a}})Hendrycks, Burns, Basart, Critch, Li, Song, and Steinhardt}]{hendrycks2021ethics}
Dan Hendrycks, Collin Burns, Steven Basart, Andrew Critch, Jerry Li, Dawn Song, and Jacob Steinhardt. 2021{\natexlab{a}}.
\newblock Aligning ai with shared human values.
\newblock \emph{Proceedings of the International Conference on Learning Representations (ICLR)}.

\bibitem[{Hendrycks et~al.(2021{\natexlab{b}})Hendrycks, Burns, Basart, Zou, Mazeika, Song, and Steinhardt}]{hendryckstest2021}
Dan Hendrycks, Collin Burns, Steven Basart, Andy Zou, Mantas Mazeika, Dawn Song, and Jacob Steinhardt. 2021{\natexlab{b}}.
\newblock Measuring massive multitask language understanding.
\newblock \emph{Proceedings of the International Conference on Learning Representations (ICLR)}.

\bibitem[{Hu et~al.(2021)Hu, Shen, Wallis, Allen-Zhu, Li, Wang, Wang, and Chen}]{hu2021lora}
Edward~J Hu, Yelong Shen, Phillip Wallis, Zeyuan Allen-Zhu, Yuanzhi Li, Shean Wang, Lu~Wang, and Weizhu Chen. 2021.
\newblock Lora: Low-rank adaptation of large language models.
\newblock \emph{arXiv preprint arXiv:2106.09685}.

\bibitem[{Jiang et~al.(2023)Jiang, Sablayrolles, Mensch, Bamford, Chaplot, Casas, Bressand, Lengyel, Lample, Saulnier et~al.}]{jiang2023mistral}
Albert~Q Jiang, Alexandre Sablayrolles, Arthur Mensch, Chris Bamford, Devendra~Singh Chaplot, Diego de~las Casas, Florian Bressand, Gianna Lengyel, Guillaume Lample, Lucile Saulnier, et~al. 2023.
\newblock Mistral 7b.
\newblock \emph{arXiv preprint arXiv:2310.06825}.

\bibitem[{Karpathy(2022)}]{NanoGPT}
Andrej Karpathy. 2022.
\newblock Nanogpt.
\newblock \url{https://github.com/karpathy/nanoGPT}.

\bibitem[{Kim et~al.(2021)Kim, Hong, Kim, Kang, and Myaeng}]{kim-etal-2021-seen}
Jeonghwan Kim, Giwon Hong, Kyung-min Kim, Junmo Kang, and Sung-Hyon Myaeng. 2021.
\newblock \href {https://doi.org/10.18653/v1/2021.emnlp-main.563} {Have you seen that number? investigating extrapolation in question answering models}.
\newblock In \emph{Proceedings of the 2021 Conference on Empirical Methods in Natural Language Processing}, pages 7031--7037, Online and Punta Cana, Dominican Republic. Association for Computational Linguistics.

\bibitem[{Lee et~al.(2024)Lee, Sreenivasan, Lee, Lee, and Papailiopoulos}]{lee2024teaching}
Nayoung Lee, Kartik Sreenivasan, Jason~D. Lee, Kangwook Lee, and Dimitris Papailiopoulos. 2024.
\newblock \href {https://openreview.net/forum?id=dsUB4bst9S} {Teaching arithmetic to small transformers}.
\newblock In \emph{The Twelfth International Conference on Learning Representations}.

\bibitem[{Penedo et~al.(2023)Penedo, Malartic, Hesslow, Cojocaru, Cappelli, Alobeidli, Pannier, Almazrouei, and Launay}]{refinedweb}
Guilherme Penedo, Quentin Malartic, Daniel Hesslow, Ruxandra Cojocaru, Alessandro Cappelli, Hamza Alobeidli, Baptiste Pannier, Ebtesam Almazrouei, and Julien Launay. 2023.
\newblock \href {http://arxiv.org/abs/2306.01116} {The {R}efined{W}eb dataset for {F}alcon {LLM}: outperforming curated corpora with web data, and web data only}.
\newblock \emph{arXiv preprint arXiv:2306.01116}.

\bibitem[{Pfau et~al.(2024)Pfau, Merrill, and Bowman}]{pfau2024let}
Jacob Pfau, William Merrill, and Samuel~R Bowman. 2024.
\newblock Let's think dot by dot: Hidden computation in transformer language models.
\newblock \emph{arXiv preprint arXiv:2404.15758}.

\bibitem[{Sennrich et~al.(2015)Sennrich, Haddow, and Birch}]{sennrich2015neural}
Rico Sennrich, Barry Haddow, and Alexandra Birch. 2015.
\newblock Neural machine translation of rare words with subword units.
\newblock \emph{arXiv preprint arXiv:1508.07909}.

\bibitem[{Shen et~al.(2023)Shen, Bubeck, Eldan, Lee, Li, and Zhang}]{shen2023positional}
Ruoqi Shen, S{\'e}bastien Bubeck, Ronen Eldan, Yin~Tat Lee, Yuanzhi Li, and Yi~Zhang. 2023.
\newblock Positional description matters for transformers arithmetic.
\newblock \emph{arXiv preprint arXiv:2311.14737}.

\bibitem[{Singh and Strouse(2024)}]{singh2024tokenization}
Aaditya~K Singh and DJ~Strouse. 2024.
\newblock Tokenization counts: the impact of tokenization on arithmetic in frontier llms.
\newblock \emph{arXiv preprint arXiv:2402.14903}.

\bibitem[{Touvron et~al.(2023)Touvron, Lavril, Izacard, Martinet, Lachaux, Lacroix, Rozi{\`e}re, Goyal, Hambro, Azhar et~al.}]{touvron2023llama}
Hugo Touvron, Thibaut Lavril, Gautier Izacard, Xavier Martinet, Marie-Anne Lachaux, Timoth{\'e}e Lacroix, Baptiste Rozi{\`e}re, Naman Goyal, Eric Hambro, Faisal Azhar, et~al. 2023.
\newblock Llama: Open and efficient foundation language models.
\newblock \emph{arXiv preprint arXiv:2302.13971}.

\bibitem[{Warstadt et~al.(2023)Warstadt, Mueller, Choshen, Wilcox, Zhuang, Ciro, Mosquera, Paranjabe, Williams, Linzen et~al.}]{warstadt2023findings}
Alex Warstadt, Aaron Mueller, Leshem Choshen, Ethan Wilcox, Chengxu Zhuang, Juan Ciro, Rafael Mosquera, Bhargavi Paranjabe, Adina Williams, Tal Linzen, et~al. 2023.
\newblock Findings of the babylm challenge: Sample-efficient pretraining on developmentally plausible corpora.
\newblock In \emph{Proceedings of the BabyLM Challenge at the 27th Conference on Computational Natural Language Learning}.

\bibitem[{Wei et~al.(2022)Wei, Wang, Schuurmans, Bosma, Xia, Chi, Le, Zhou et~al.}]{wei2022chain}
Jason Wei, Xuezhi Wang, Dale Schuurmans, Maarten Bosma, Fei Xia, Ed~Chi, Quoc~V Le, Denny Zhou, et~al. 2022.
\newblock Chain-of-thought prompting elicits reasoning in large language models.
\newblock \emph{Advances in neural information processing systems}, 35:24824--24837.

\end{thebibliography}
\bibliographystyle{acl_natbib}



\end{document}